\documentclass{article}
\usepackage{arxiv}

\usepackage[utf8]{inputenc} 
\usepackage[T1]{fontenc}    
\usepackage[hidelinks]{hyperref}       
\usepackage{url}            
\usepackage{booktabs}       
\usepackage{amsfonts}       
\usepackage{nicefrac}       
\usepackage{microtype}      
\usepackage{lipsum}		
\usepackage{graphicx}
\usepackage{natbib}
\usepackage{doi}

\usepackage{amsmath,amssymb,amsfonts}
\usepackage{amsthm}
\usepackage{mathrsfs}%
\usepackage{subcaption}
\usepackage{graphicx}%
\usepackage{tabularx,booktabs,siunitx}
\usepackage{array}
\usepackage{multicol}
\usepackage{multirow}
\usepackage{tabularx}
\usepackage{color, colortbl}
\usepackage{makecell}
\usepackage{subcaption}
\usepackage{comment}
\usepackage{xurl}  
\usepackage{lineno}

\newcolumntype{L}[1]{>{\raggedright\arraybackslash}m{#1}}
\newcolumntype{C}[1]{>{\centering\arraybackslash}m{#1}}
\newcolumntype{R}[1]{>{\raggedleft\arraybackslash}m{#1}}
\newcommand{\celdat}[1]{\normalsize\bfseries #1}

\newcommand{\celdah}[1]{\footnotesize #1}
\newcommand{\celda}[1]{\footnotesize\bfseries #1}

\title{Adaptive Gaussian Mixture Models-Based Anomaly Detection for Under-Constrained Cable-Driven Parallel Robots}

\author{ \href{https://orcid.org/0000-0001-9974-9465}{\includegraphics[scale=0.06]{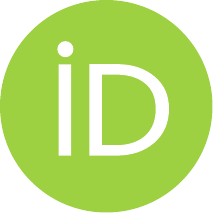}\hspace{1mm}Julio Garrido}\\
	Dept. of Automation and Systems Engineering\\
	Universidade de Vigo\\
	Vigo, 36310 Pontevedra \\
	\texttt{jgarri@uvigo.gal} \\
	\And
	\href{https://orcid.org/0000-0002-8761-7235}{\includegraphics[scale=0.06]{orcid.pdf}\hspace{1mm}Javier Vales} \\
	Dept. of Communication and Information Technologies\\
	Technical University of Cartagena \\
	Cartagena, 30202 Murcia\\
	\texttt{javier.vales@upct.es} \\
	\And
	\href{https://orcid.org/0000-0002-2233-8606}{\includegraphics[scale=0.06]{orcid.pdf}\hspace{1mm}Diego Silva-Muñiz}\\
	Dept. of Automation and Systems Engineering\\
	Universidade de Vigo\\
	Vigo, 36310 Pontevedra \\
	\texttt{diego.silva.muniz@uvigo.gal} \\
	\And
	\href{https://orcid.org/0000-0002-0153-9310}{\includegraphics[scale=0.06]{orcid.pdf}\hspace{1mm}Enrique Riveiro}\\
	Dept. of Automation and Systems Engineering\\
	Universidade de Vigo\\
	Vigo, 36310 Pontevedra \\
	\texttt{enrriveiro@uvigo.gal} \\
	\And
	\href{https://orcid.org/0000-0002-9881-6192}{\includegraphics[scale=0.06]{orcid.pdf}\hspace{1mm}Pablo López-Matencio} \\
	Dept. of Communication and Information Technologies\\
	Technical University of Cartagena\\
	Cartagena, 30202 Murcia\\
	\texttt{pablo.lopez@upct.es} \\
	\And
	\href{https://orcid.org/0009-0009-3591-7154}{\includegraphics[scale=0.06]{orcid.pdf}\hspace{1mm}Josué Rivera-Andrade}\\
	Dept. of Automation and Systems Engineering\\
	Universidade de Vigo\\
	Vigo, 36310 Pontevedra \\
	\texttt{josueroberto.rivera.andrade@uvigo.gal} \\
}

\renewcommand{\shorttitle}{J. Garrido et al.}

\hypersetup{
pdftitle={Adaptive Gaussian Mixture Models-based anomaly detection for under-contrained cable-driven parallel robots},
pdfsubject={cs.RO, cs.AI},
pdfauthor={J. Garrido, J. Vales, D. Silva-Muñiz, E. Riveiro, P. López-Mantencio, J. Rivera-Andrade},
pdfkeywords={Anomaly detection, Cable-Driven Parallel Robots, Gaussian Mixture Models, Robotic Control Systems, Unsupervised Learning},
}

\begin{document}
\maketitle

\begin{abstract}
 Cable-Driven Parallel Robots (CDPRs) are increasingly used for load manipulation tasks involving predefined toolpaths with intermediate stops. At each stop, where the platform maintains a fixed pose and the motors keep the cables under tension, the system must evaluate whether it is safe to proceed by detecting anomalies that could compromise performance (e.g., wind gusts or cable impacts). This paper investigates whether anomalies can be detected using only motor torque data, without additional sensors. It introduces an adaptive, unsupervised outlier detection algorithm based on Gaussian Mixture Models (GMMs) to identify anomalies from torque signals. The method starts with a brief calibration period—just a few seconds—during which a GMM is fit on known anomaly-free data. Real-time torque measurements are then evaluated using Mahalanobis distance from the GMM, with statistically derived thresholds triggering anomaly flags. Model parameters are periodically updated using the latest segments identified as anomaly-free to adapt to changing conditions. Validation includes 14 long-duration test sessions simulating varied wind intensities. The proposed method achieves a 100\% true positive rate and 95.4\% average true negative rate, with 1-second detection latency. Comparative evaluation against power threshold and non-adaptive GMM methods indicates higher robustness to drift and environmental variation.
\end{abstract}


\section{Introduction}\label{sec1}
Cable-driven parallel robots (CDPRs) are robotic systems where the
end-effector's position and orientation are controlled by adjusting the length
of multiple cables, offering advantages in cost, performance, and adaptability
\cite{zarebidokiReview2022}. Their structure, composed of cables and winches,
enables larger workspaces and higher payload capacities compared to
anthropomorphic robots \cite{martin-parraNovel2024}. Unlike other parallel
robots like Delta or Stewart platforms, CDPRs are also more reconfigurable and
capable of complex motions, making them ideal for applications that require
extended workspaces and high load capacities \cite{bobyMeasurement2021,
	izardCable2023}.  

However, their reliance on cables introduces a key limitation: cables can only
operate under tension, not compression. As a result, stability depends on the
specific setup of the systems. For instance, an eight-cable setup (four upper
and four lower) provides precise control and robustness, while a four-cable
configuration, though suitable for larger workspaces, is under-constrained and
more prone to oscillations in the end-effector \cite{korayemDynamics2017}.

Research has explored CDPR applications in diverse fields such as 3D printing
\cite{qianCalibration2024}, rehabilitation \cite{huoDevelopment2024}, storage
and retrieval systems \cite{bruckmannDevelopment2012}, sports tracking devices \cite{ghanatianExperimental2023a}, precision
agriculture \cite{prabhaCable2021} and load manipulation \cite{cullaFull2018, kimRemotely2022}.  Despite the aforementioned advantages, their
industrial adoption remains limited
\cite{metillonPerformance2022,boumannDevelopment2019}. The main barriers to
widespread implementation are the scarcity of commercial solutions, with only a
few available, such as RBOT9 and CRANEBOT \cite{rbot9automationROCAP,
	CRANEBOT}, and challenges in meeting industrial safety and operational standards required
for seamless integration with industrial controllers.

Enhancing safety could significantly boost CDPR adoption, but research has
mainly focused on cable breakage and collisions. This kind of solutions include
adding passive auxiliary cables with clamping devices for redundancy
\cite{caroFailure2020} to reconfigurable architectures that actively relocate cable anchor points for task recovery after a cable failure \cite{ramanFailure2022}, as well as collision-management techniques that address external impacts with humans or the environment \cite{rousseauHumancable2022}, and internal self-interference \cite{blanchetInterference2014}.

Besides, as stated, under-constrained CDPRs are more sensitive
to external forces like wind, causing lateral disturbances that impact static
positioning in pick-and-place tasks \cite{husseinGeometric2018}. Accurate
endpoint positioning is crucial, while trajectory errors during motion are more
tolerable. Deviations of the end-effector due to external factors like wind or
unintended operator contact are anomalies of the expected and safe behavior
\cite{taghaviDevelopment2018}. These anomalies can compromise grip quality or
degrade outcomes in applications like laser inspection. Therefore, detecting
these anomalies is the first step toward implementing actions to mitigate them.

Although machine learning, especially deep learning, has improved anomaly
detection \cite{choiDeep2021, pangDeep2022}, many solutions are highly specialized and
require extensive domain knowledge. Traditional methods like k-Nearest Neighbour,
Decision Trees, or Discriminant Analysis have been applied to detect cable
breakage or sensor failures in CDPRs \cite{bettegaLoad2024}, and Neural
Networks have identified deviations from standard toolpaths
\cite{jabbariaslAdaptive2017a}. Despite recent advances, CDPRs still lack practical anomaly detection strategies that can learn on the fly, without pre-labeled anomalies, and remain robust to wind gusts.

We present an approach to detect unintended end-effector movements by
analyzing anomalies in cable tensions without requiring additional external
sensors or detailed environmental information. Our study focuses on conditions in which motion control maintains the robot's pose while the motors remain powered and the motor brakes are switched off. These conditions occur, for instance, when the manipulator remains at rest while performing a picking or placing operation with its tool, as in typical pick-and-place tasks. We
investigate whether disturbances from wind or other external forces can be
detected solely through motor torque measurements based on servo amplifier
current consumption. By doing so, any detected anomalies can prompt the
postponement of subsequent operations until the anomaly is resolved. This
approach enhances safety and provides redundancy for critical decisions.  From an implementation perspective, cable tension are inferred directly from motor torques, avoiding the extra hardware, wiring, and calibration effort required by cable-mounted load cells as proposed in \cite{gaoDetection2024}.

The rest of this paper is organized as follows. Section~\ref{STABILITY} analyzes the stability conditions that allow the application of the anomaly detection method described in Section~\ref{ANOMALIES} for under-constrained CDPRs. Section~\ref{SETUP} describes the experimental setup used and the experiments conducted. Section~\ref{RESULTS} presents the results obtained from the various rounds of experiments. Finally, Section~\ref{CONCLUSIONS} summarizes the conclusions. All sources and csv datafiles are available at GitHub: \url{https://github.com/javiervales/cdpr}.

\section{Static Stability of Under-Constrained Cable-Driven Parallel Robots}
\label{STABILITY}
Stability describes the response of a mechanical system to a disturbance and is a fundamental property in the design and operation of CDPRs, particularly for under-constrained systems like our 4-cable experimental system (see Section~\ref{SETUP}), where gravity significantly influences performance \cite{pottCableDriven2018}. A configuration is considered \emph{stable} when a slight change in the cable forces does not permanently displace the platform or prevent it from returning to its original position once the disturbance disappears. Previous research has addressed the stability of CDPRs through analytical methods that involve the use of Lagrange multipliers and the computation of the Hessian matrix, as in \cite{carricatoGeometricoStatic2010} and \cite{abbasnejadDirect2015}, respectively. Specifically, following these works, stability is assured when the reduced Hessian ($\mathbf{H}_{\mathrm{r}}$) is positive definite.

We define an \emph{equilibrium configuration} as any pose in which the cable forces exactly counter-balance the external load (only gravity on the platform’s center of mass is considered) so that the resultant force and moment on the platform are both zero. When, in addition, the platform is at rest, the configuration becomes a \emph{static equilibrium}. This study focuses exclusively on such static equilibrium and examines whether the CDPR remain stable under infinitesimal tension perturbations, a property hereafter referred to as \emph{static stability}. Establishing the system's static stability 
is mandatory because Gaussian Mixture Models (GMMs) cannot distinguish intrinsic variations from external outliers, thus it is required that internal variations do
not exists, or at least being moderated.

\begin{figure}
	\centering
	\includegraphics[width=0.425\columnwidth]{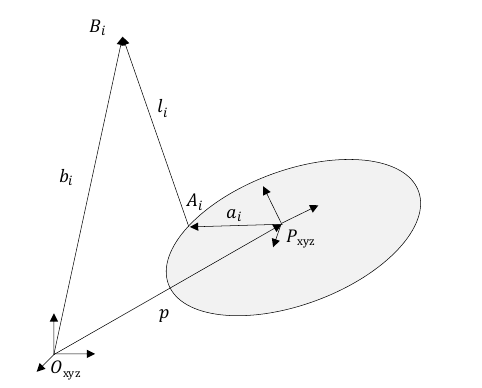}
	\caption{Closed-loop schematic of one cable in a CDPR.}
	\label{fig:cdpr_vectors}
\end{figure}

The CDPR model consists of a mobile platform connected to the base by $n$ inelastic, massless cables, as shown in Figure \ref{fig:cdpr_vectors}. $O_{\mathrm{xyz}}$ represents the system's coordinate frame and $P_{\mathrm{xyz}}$ is the mobile coordinate frame of the platform. The position of the platform is defined by the position vector $\boldsymbol{p}$ and the rotation matrix $\mathbf{R}$, parameterized by Euler angles $\varepsilon=[\varphi, \psi, \theta]^{\mathrm{T}}$, with generalized coordinates $\boldsymbol{q} = [\boldsymbol{p}^\mathrm{T}, \varepsilon^\mathrm{T}]^\mathrm{T}$. $B_i$ denotes the distal anchor points where the cables originate, while $A_i$ are the proximal anchor points where the cables attach to the mobile platform. The $i$-th cable is represented by vector $\boldsymbol{l}_i$ (Equation~\eqref{eq:li}).

To simplify the model, we neglect the influence of pulley transmission on platform accelerations, cable tensions, and workspace boundaries. This assumption is supported by prior studies showing that omitting pulley effects reduces the workspace volume by approximately 1\% when pulley radii are around 2.5\% of the robot's shortest edge \cite{pottInfluence2012}. Although pulley kinematics may temporarily amplify platform acceleration by up to 50\% in worst-case scenarios \cite{idaRestRest2019}, such transient effects are typically mitigated by the inner control loops of the drives and do not significantly impact the steady-state behavior under analysis.

\begin{equation}\label{eq:li}
	\boldsymbol{l}_i = B_i - A_i = \boldsymbol{b}_i - \boldsymbol{p} - \boldsymbol{r}_i = \boldsymbol{b}_i - \boldsymbol{p} - \mathbf{R} \boldsymbol{a}_i
\end{equation}

Using the notation introduced by \cite{carricatoGeometricoStatic2010}, the static stability condition reads

\begin{equation}\label{eq:equilibrium}
	\sum_{i=1}^{n} \frac{\tau_i}{\|\boldsymbol{l}_i\|} \, \boldsymbol{\$}_i + Q \, \boldsymbol{\$}_e = 0, \quad \tau_i \geq 0, \quad i = 1, \dots, n
\end{equation}

where $(\tau_i / \|\boldsymbol{l}_i\|) \boldsymbol{\$}_i$ is the force applied to the $i$-th cable, and \(Q \boldsymbol{\$}_e\) is the external wrench. 

Under this static stability assumption, friction dissipates kinetic energy, leading the platform to a state of equilibrium over time. This result stems from the Lagrangian function \(\mathcal{L}\), which relates the kinetic and potential energy to the system's kinematics and dynamics. Following the criterion in \cite{carricatoGeometricoStatic2010}, the reduced Hessian
$\mathbf{H_r}$ of $\mathcal{L}$ must be positive definite for the equilibrium defined in
Equation~\eqref{eq:equilibrium}. Its expression is

\begin{equation}\label{eq:hr}
	\mathbf{H}_r = \mathbf{N}_p^{\mathrm{T}} \, \mathbf{H}_p \, \mathbf{N}_p
\end{equation}

\begin{equation}\label{eq:hp}
	\mathbf{H}_p = \sum_{i=1}^n \frac{\tau_i}{\|\boldsymbol{l}_i\|} 
	\begin{bmatrix}
		\mathrm{I} & -\tilde{\boldsymbol{r}}_i \\[4pt]
		\tilde{\boldsymbol{r}}_i & \dfrac{1}{2} \left( 
		\tilde{\boldsymbol{r}}_i \tilde{\boldsymbol{p}} - \tilde{\boldsymbol{r}}_i \tilde{\boldsymbol{b}}_i + \tilde{\boldsymbol{p}} \tilde{\boldsymbol{r}}_i - \tilde{\boldsymbol{b}}_i \tilde{\boldsymbol{r}}_i 
		\right)
	\end{bmatrix}
\end{equation}

\begin{equation}\label{eq:jb}
	\mathbf{J}_p = 
	\begin{bmatrix}
		\boldsymbol{l}_1^{\mathrm{T}} & (\boldsymbol{r}_1 \times \boldsymbol{l}_1)^{\mathrm{T}} \\[4pt]
		\vdots & \vdots \\[4pt]
		\boldsymbol{l}_n^{\mathrm{T}} & (\boldsymbol{r}_n \times \boldsymbol{l}_n)^{\mathrm{T}}
	\end{bmatrix}
\end{equation}

where $\mathbf{N}_p$ denotes the kernel (or null space) of the homomorphism $\mathbf{J}_p$ (Equation~\eqref{eq:jb}) and $\tilde{\boldsymbol{r}}$ is the skew-symmetric matrix of dimension $n$. This methodology will be used in Section~\ref{study} to validate the static stability of the CDPR in the pose of the experiment.

\section{Experimental setup}
\label{SETUP}
\begin{figure}[t!]
	\centering
	\begin{subfigure}[b]{0.525\textwidth}
		\centering
		\includegraphics[width=\textwidth]{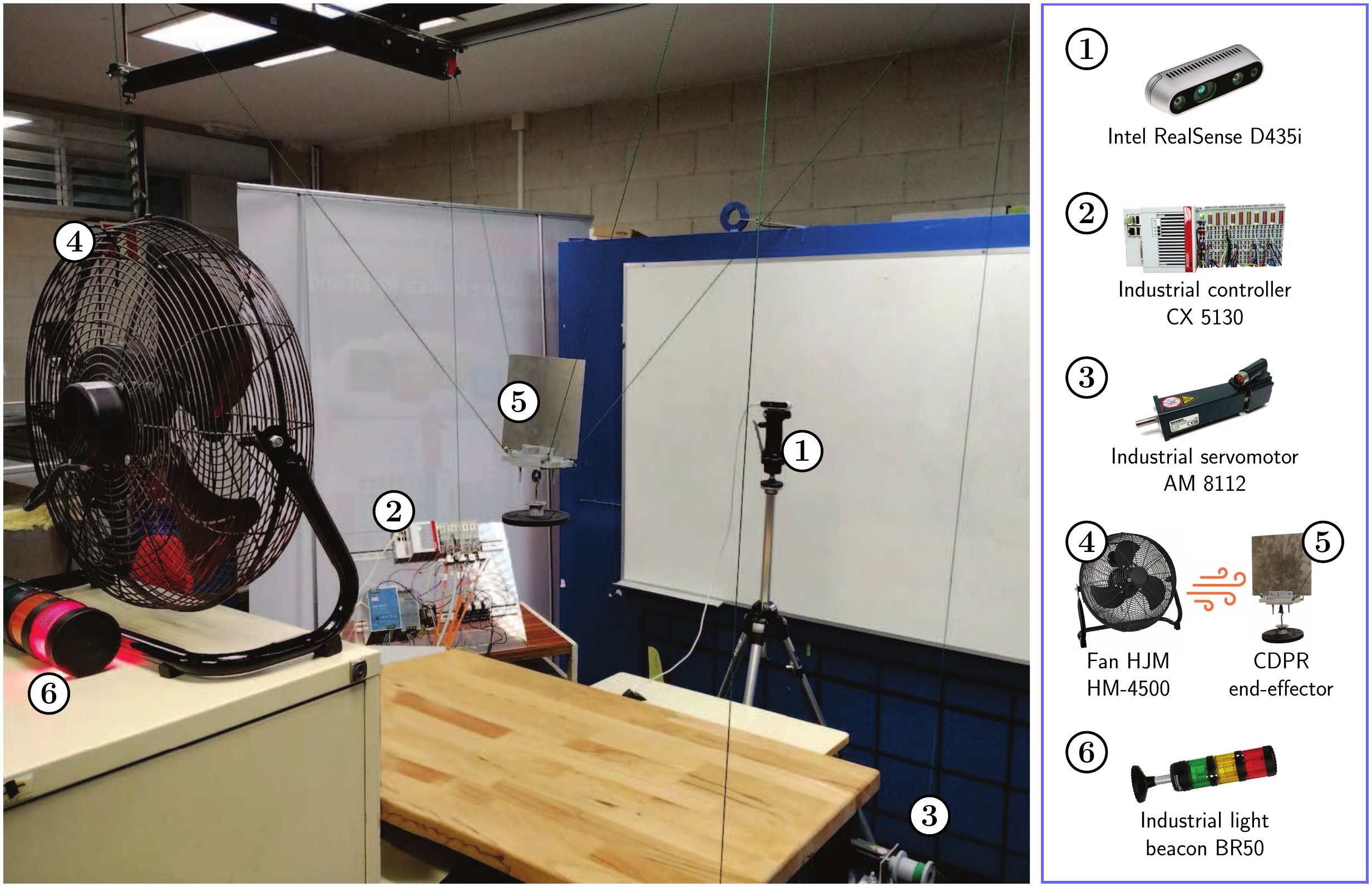}
		\caption{}
		\label{CDPRSETUP}
	\end{subfigure}
	\hfill
	\begin{subfigure}[b]{0.4525\textwidth}
		\centering
		\includegraphics[width=\textwidth]{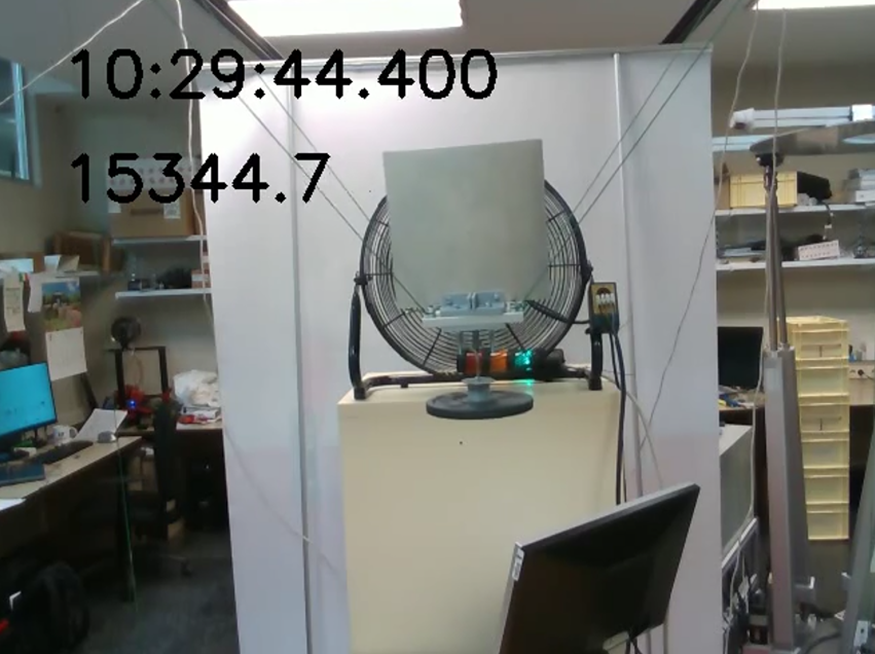}
		\caption{}
		\label{fig:cdpr_system_b}
	\end{subfigure}
	\caption{Experimental setup for CDPR testing under wind disturbances. (a) Photograph of the experiment. (b) Capture of the video recorded during the tests with the timestamp data.}
	\label{fig:cdpr_system}
\end{figure}

An experimental setup was designed to evaluate the anomaly-detection system (detailed in Section~\ref{ANOMALIES}) through controlled experiments in which disturbances simulate external forces and environmental interference similar to those in actual facilities. 
The setup, shown in Figure~\ref{CDPRSETUP}, consists of a 4-cable CDPR and a fan with four
adjustable strength levels (0 for off to 3 for maximum) to simulate external outliers (item 4 in Figure~\ref{CDPRSETUP}). A tri-color beacon (item 6 in Figure~\ref{CDPRSETUP}) provides an visual feedback of the current fan setting in the recorded images, as shown in the center of Figure~\ref{fig:cdpr_system_b}.

The mechanical prototype of the CDPR comprised several components. The dimensions of the structure are 860$\times$1420 (length and width respectively)  at a height of 1600 mm and utilizes aluminum profiles with a 40$\times$40 mm cross-section. At the extreme ends of this structure are the polyurethane pulleys with a 25 mm radius. The cables are made of Dyneema SK75 with a diameter of 1.5 mm,
while the winches, fabricated from PLA material, have a radius of 45 mm. The
aluminum end-effector measures 66$\times$114\-$\times$18 mm and weighs 0.602
kg. To increase its surface area and improve its exposure to airflow, an
aluminum sheet measuring 195$\times$195$\times$1 mm is attached to the end-effector as a ``sail''. A weight of 1 kg is suspended from the end effector to ensure significant tension on the cables. 

Inside the controller, kinematics, trajectory generation and control structures are implemented in Structured Text (IEC 61131-3 \cite{iecProgrammable2013}) under the TwinCAT 3 runtime \cite{beckhoffautomationgmbh&co.kgTwinCAT}, with motor control compliant with PLCopen Motion Control standards \cite{plcopenFunction2008}. Torque and position data is forwarded over EtherCAT to a digital I/O card and then to four EL7211 servo drives. Each servo drive powers an AM8112 servomotor that reels the cables onto the winch.

\begin{figure}
	\centering
	\includegraphics[width=\textwidth]{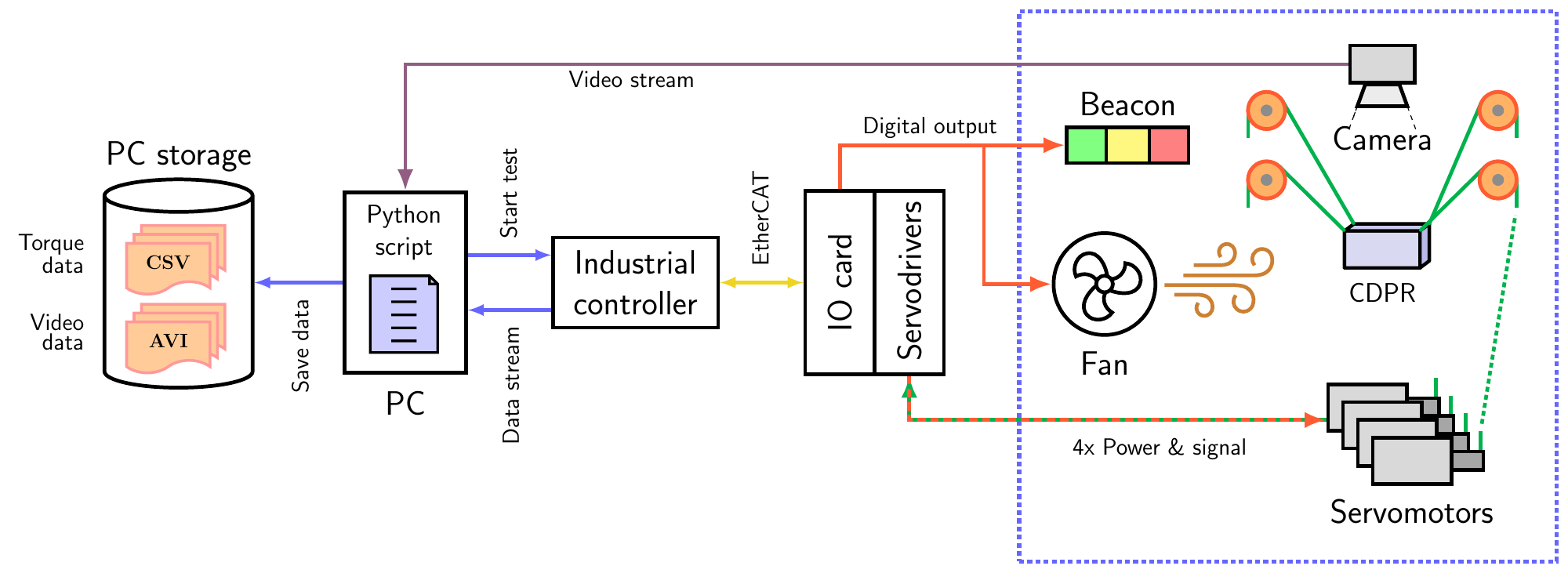}
	\caption{Signal schematic and component interaction for data acquisition and control of the CDPR under external disturbances.}
	\label{fig:cdpr_arch}
\end{figure}

The signal-and-data flow, used for the operation of the CDPR, is detailed in the architecture diagram of Figure~\ref{fig:cdpr_arch}. A script running on a PC launches each trial, issues starting commands to the Beckhoff CX5130 industrial controller (item 2 in Figure~\ref{CDPRSETUP}), and receives a continuous stream of internal variables (cable tensions, platform position, and timestamps). The script also opens a data file for the incoming motor torque stream and starts recording video footage from the Intel RealSense depth camera (item 1 in Figure~\ref{CDPRSETUP}). Both video and data file are timestamped for synchronization purposes. A view from the recorded video is provided in Figure~\ref{fig:cdpr_system_b}.

During the tests, the CDPR remained in a stable position (see next section~\ref{study}) with consistent airflow impacting it always from the
same direction when an anomaly takes place. The end-effector position was fixed at
coordinates (0, 0, 700) mm. It is important to note that all tests were
conducted only after the motors reached a stable operating temperature. The
experiments were carried out in a laboratory testbed with temperature and
humidity controlled at approximately 20 \textsuperscript{o}C and 75\%,
respectively. For each test, motor torques (as proxies for cable
tensions) are timestamped and saved at a rate of 100 samples/s.

The tests conducted are as follows:

\begin{enumerate}
	\item \textbf{Gust test (Type 1)}: The fan alternated between being off and
	operating at the selected wind strength for a given period of time (usually 10
	min). These tests established a baseline for the anomaly detection algorithm,
	enabling it to accurately classify the presence or absence of wind.
	
	\item \textbf{Rest and gust cycle test (Type 2)}: Following a prolonged resting
	period (fan off) of several hours, the system underwent 2 hours of short gust
	sessions with 10-minute on-off cycles. These tests evaluated the algorithm's
	ability to handle motor torque drifts, defined as the gradual variation in delivered torque that arises when the motor current changes over time.
	
	\item \textbf{Random wind disturbance test (Type 3)}: After an initial resting
	period, variable disturbances were applied, varying in strength (levels 0 to 3) and duration (1 to 20 minutes). These tests are aimed to assess the algorithm's ability to detect outliers under
	dynamic wind conditions.
\end{enumerate}

\subsection{Study of static stability for the 4-cable CDPR}\label{study}
The stability study presented next for our CDPR is intended to confirm that the anomaly-detection system described in Section~\ref{ANOMALIES} can be applied to this model. 

The structure and end-effector dimensions used in Eqs.~\ref{eq:hp} and \ref{eq:jb} are those given in the previous Section~\ref{SETUP}, and the position and orientation of the end-effector are $\boldsymbol{p}$=$[0 \; 0 \; 0.7]^{\mathrm{T}}$ m and $\varepsilon$=$[0 \; 0 \; 0]^T$ rad, respectively. The cable tensions are $\boldsymbol{\tau}$=$[15.89 \; 11.19 \; 15.2 \; 12.57]$ N.
The cable vectors $\boldsymbol{l}_i$ are: $\boldsymbol{l}_1$=$[-0.653 \; -0.397 \; 0.9]^{\mathrm{T}}$, $\boldsymbol{l}_2$=$[0.653
\; -0.397 \; 0.9]^{\mathrm{T}}$, $\boldsymbol{l}_3$=$[0.653 \; 0.397 \; 0.9]^{\mathrm{T}}$, and $\boldsymbol{l}_4$=$[-0.653 \;
0.397 \; 0.9]^{\mathrm{T}}$ m. All cables have length 1.1807 m. Evaluating $\mathbf{H}_r$ in Equation~\eqref{eq:hr} leads to:

\begin{equation}
	\label{eq:hr2}
	\mathbf{H}_r = \mathbf{N}_p^{\mathrm{T}} \mathbf{H}_p \mathbf{N}_p =
	\begin{bmatrix}
		2.1530 & -0.1499 \\
		-0.1499 & 0.9146
	\end{bmatrix}
\end{equation}

Since $\mathbf{H}_r$ is symmetric and has positive eigenvalues (2.1530 and 0.9146), it is positive definite,
confirming that the CDPR is statically stable 
at the pose $\boldsymbol{q}(\boldsymbol{p}, \varepsilon)$. Consequently, any small external disturbance produces only a temporary displacement: restoring forces bring the platform back to its equilibrium position as kinetic energy dissipates. In conclusion, an outlier discriminator should theoretically be able to identify anomalies, since the CDPR is in a statically stable condition.

The next section provides a detailed explanation of the anomaly detection algorithm
developed in this study. The results from the three types of tests are
presented and thoroughly discussed in the Section~\ref{RESULTS}.
\section{Anomaly detection}
\label{ANOMALIES}

The general goal of this research is to detect anomalies using the available CDPR state
information. As discussed in Section \ref{STABILITY}, the state in an
under-constrained CDPR comprises cable lengths and motor torques. When the platform is at rest, cable lengths define the position of the end-effector, but external
forces can still alter the state of the CDPR without altering the lengths, instead
modifying only cable tensions. Consequently, our algorithm relies solely on
motor torque data.
\begin{figure*}[t!]
	\centering
	\begin{subfigure}[b]{0.49\columnwidth}
		\includegraphics[width=\linewidth]{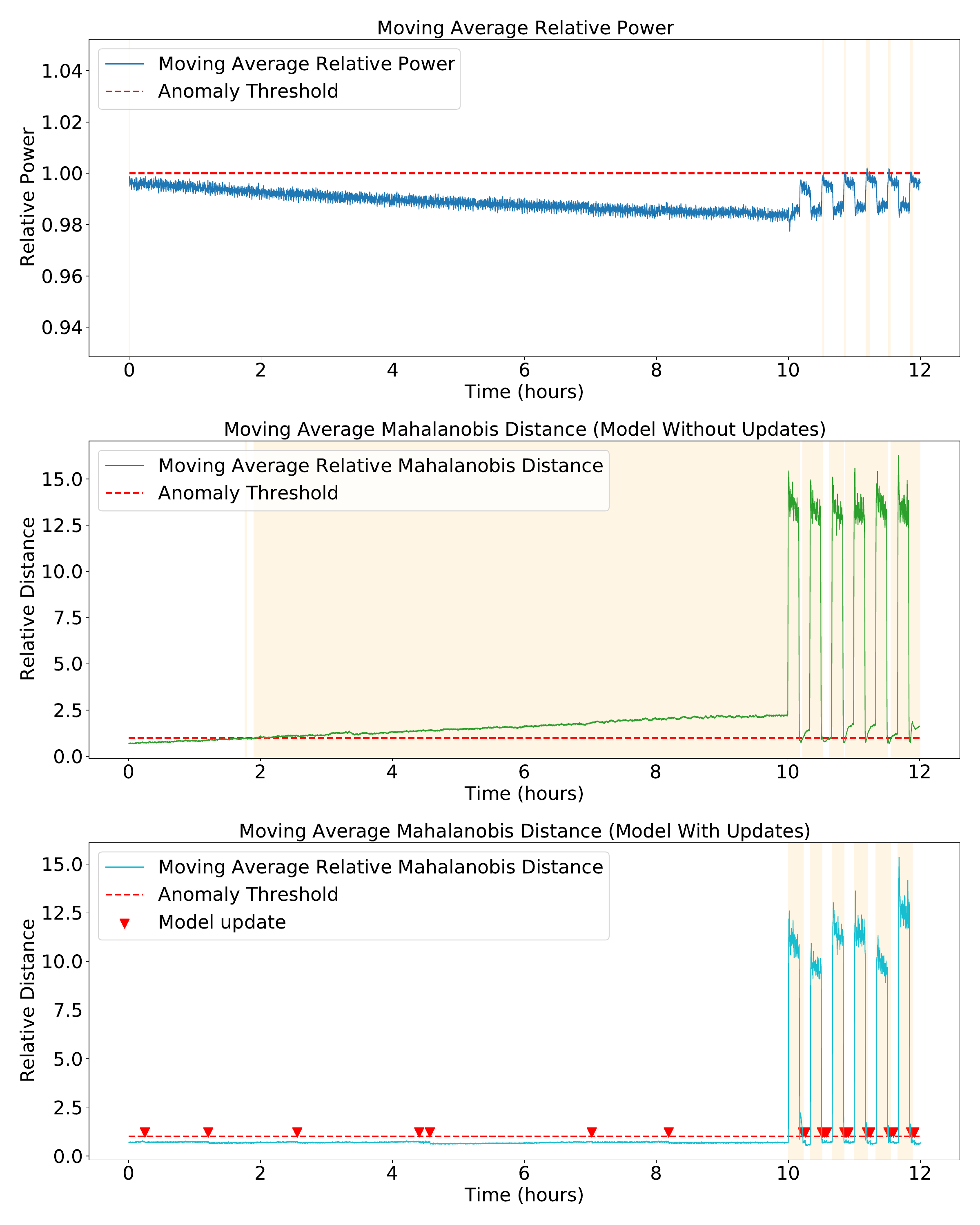}
		\caption{}
		\label{fig:test12}
	\end{subfigure}
	\hspace{0cm}
	\begin{subfigure}[b]{0.49\columnwidth}
		\includegraphics[width=\linewidth]{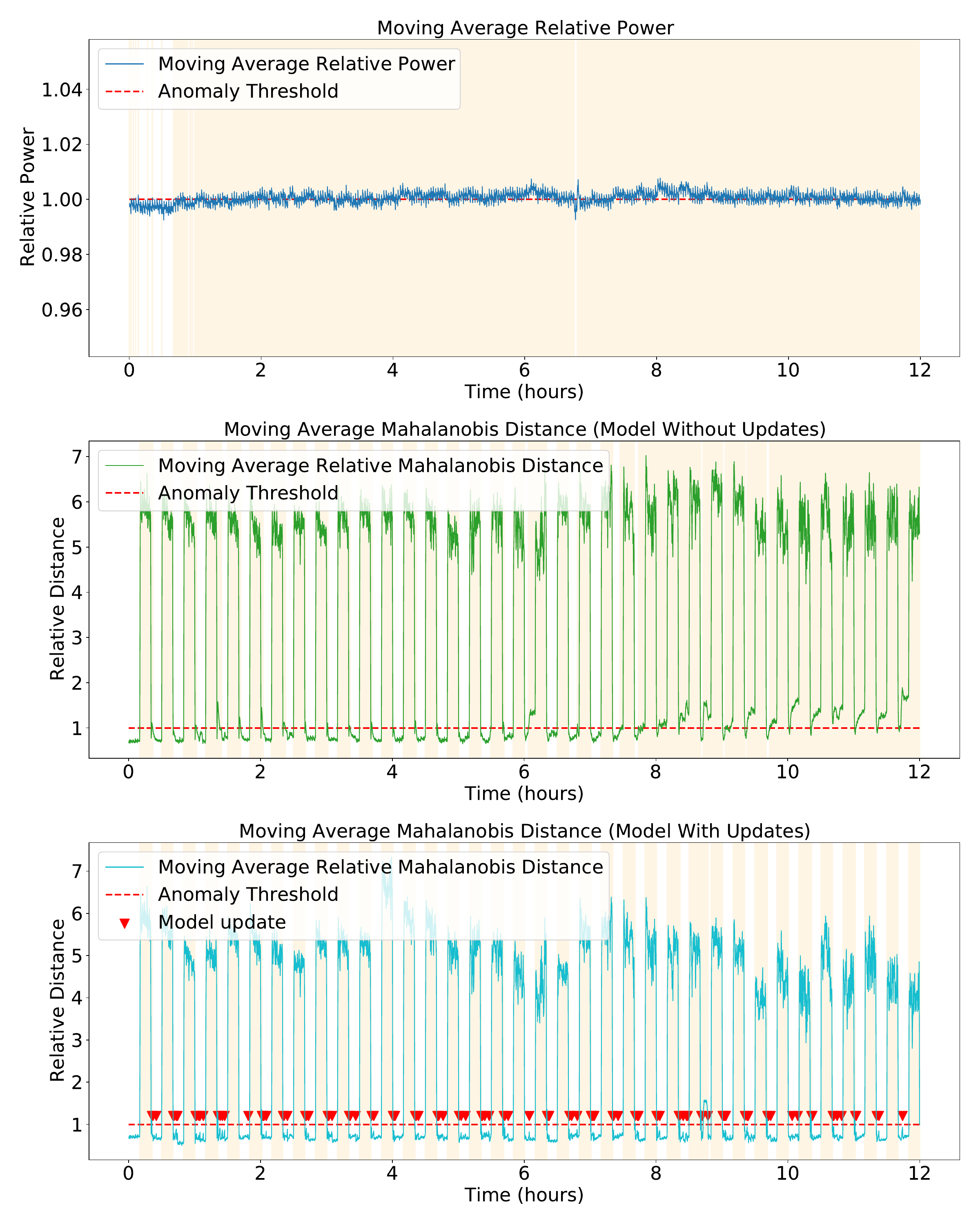}
		\caption{}
		\label{fig:test4}
	\end{subfigure}
	\caption{Relative power series and Mahalanobis distances for experiments: (a) \#12. (b) \#4 (see Table~\ref{LATABLA}).}
	\label{fig:test12and4}
\end{figure*}

An initial approach to anomaly detection could be setting a threshold for the
instantaneous detected power. However, this would underperform due to the noisy
nature of the data. For example, in the top part of Figure~\ref{fig:test12and4}, the moving-average 2-norm of the torque vector
at each time instant is shown for two of the experimental tests performed (see
Section~\ref{RESULTS}, Table~\ref{LATABLA}). As can be seen, the anomalies can be hidden within the
noise of the signal itself (Figure~\ref{fig:test4}-top). Even when
more discernible, they may still fall below the detection threshold determined
during calibration (Figure~\ref{fig:test12}-top) due to temporal
drift and the small offset from the regular signal.  Moreover, this effect of
temporal drifting has been observed in most of the experimental tests (see discussion in Section~\ref{RESULTS}). This drift in the data is not attributable to a stability
problem but to external changes such as temperature variations.
Notably, these drifts are smooth over time, suggesting that it should be
possible to detect and adapt to them.

Based on these observations, it is desirable to use a method that is less
affected by noise, achieves a greater level of distinction when an outlier
occurs, and is capable of correcting the model in the presence of temporal
drifts. To this end, we have proposed a model based on the distance
between new data and the CDPR model under stable conditions, using a
multivariate GMM. The main stages in the operation of
our algorithm are shown in Figure~\ref{OVERVIEW} and summarized next:

\begin{figure}[t]
	\centering
	\includegraphics[width=1.0\columnwidth]{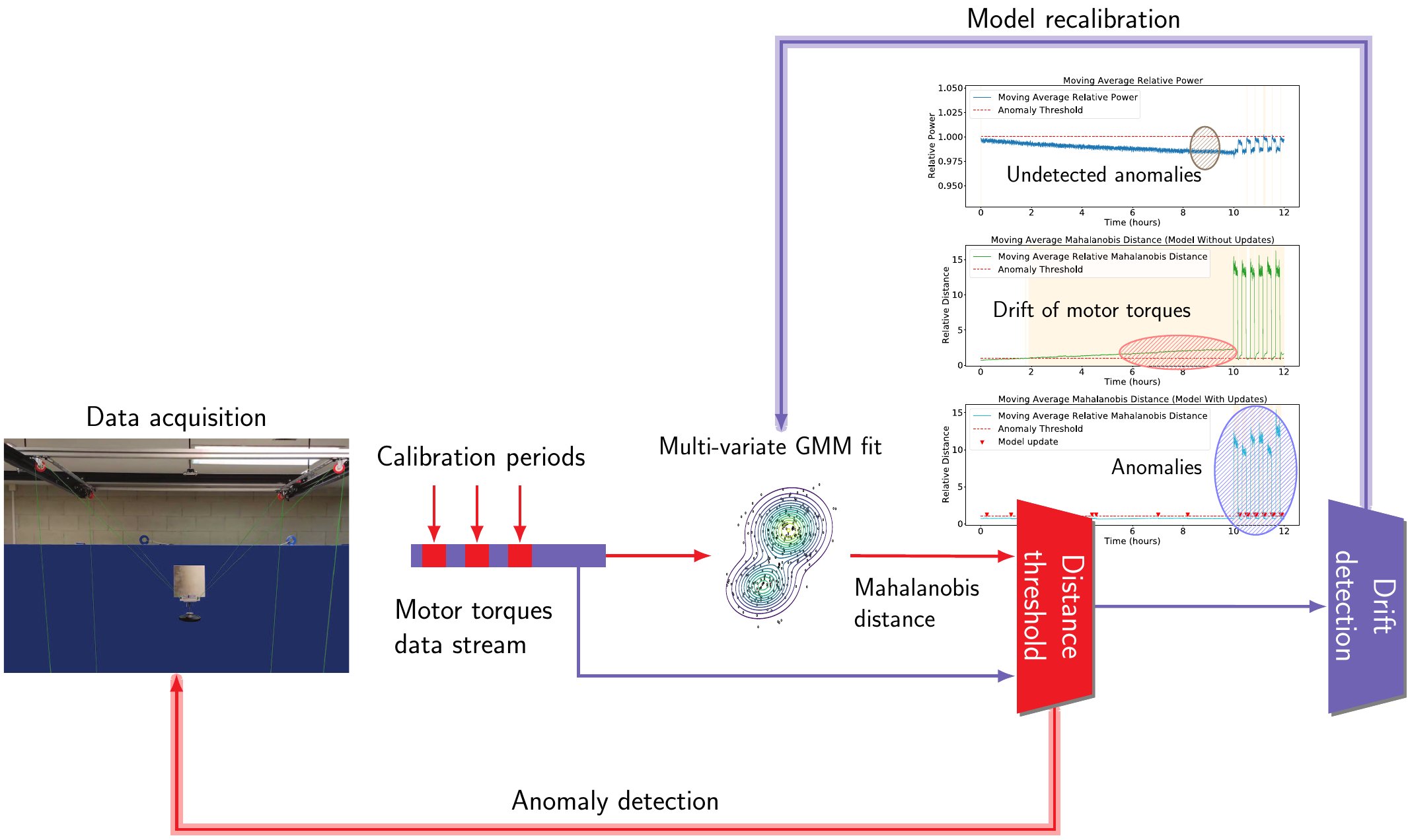}
	\caption{Model overview showing the main stages and the procedures involved in the data processing}
	\label{OVERVIEW}
\end{figure}

\begin{itemize} 
	\item \textit{Data acquisition}. At the beginning of an operation with the CDPR, baseline data is gathered while the platform is at rest and used to calibrate the model for subsequent idle periods.
	
	\item \textit{Model generation}. A multivariate GMM is fit to the calibration data. 
	
	\item \textit{Anomaly detection}. New data is compared against the GMM model
	using the Mahalanobis distance to the cluster centroids. When this distance
	exceeds the threshold, an anomaly is declared. 
	
	\item \textit{Model update}. To handle observed data drifts, we update our
	model if a significant deviation from the data is found but it is still below
	the threshold. 
\end{itemize}

Next sections describe these steps in depth.

\subsection{Data acquisition}
Torque data are read sequentially from each of the $n$ motors at each sampling
instant $j$ as an $n$-dimensional tuple $\boldsymbol{\tau}^j$.  At each processing
step, these data are grouped into sets of $M$ samples. That is, at processing
instant $j$, the data $\boldsymbol{x}^j$ is comprised by $\boldsymbol{x}^j$=$\{\boldsymbol{\tau}^j, \ldots,
\boldsymbol{\tau}^{j+M-1}\}$, thus $\boldsymbol{x}^j$ is $(n\times M)$ dimensional.

The first $C$ points $\{\boldsymbol{x}^1, \ldots, \boldsymbol{x}^C\}$ are used for
calibration. They correspond to the data recorded during the idle period at the
beginning of the experimental session, after the motors have reached their steady
temperature regime.

\subsection{Model generation}
Given the $C$ calibration points and setting the number of clusters to $K$, the GMM model
parameters (means $\boldsymbol{\mu}_k$ and covariances $\boldsymbol{\Sigma}_k$, for
$k$=$1,\ldots,K$) are fitted using the well-known Expectation-Maximization (EM)
algorithm (\emph{e.g.}, see \cite{bishopPattern2006}, Sect.~9.4.1) implemented in Python's
\texttt{scikit-learn} library through the \texttt{Gau\-ssianMixture} class. The
number of cluster $K$ is selected as the one minimizing the Bayesian
Information Criterion (BIC) for the model fit, which is commonly used in models
aimed at anomaly detection (see \cite{bishopPattern2006}, Sect.~9.4.3).

\subsection{Anomaly detection}
Once the model is fit, we can compute the distance from a new point
$\boldsymbol{x}$ to the GMM model using the Mahalanobis distance.  This is a common
measure of the novelty of a point versus a (multi-)clustered data set, which we
assume to exist since the conditions are deemed as statically stable during the
calibration phase (as studied in Section~\ref{study}). Similar approaches for the analysis of outliers in data series have been successfully applied in other domains \cite{vales-alonsoNonsupervised2023a, zhengElectromyographicBased2025, belkhoucheRobust2022}. In a multi-clustered framework the distance to the 
$k$-th cluster centroid is:

\begin{equation*}
	d_k(\boldsymbol{x}) = \left[ (\boldsymbol{x} - \boldsymbol{\mu}_k)^{\mathrm{T}} \mathbf{\Sigma}_k^{-1} (\boldsymbol{x} - \boldsymbol{\mu}_k) \right]^{\frac{1}{2}}
\end{equation*}
thus, the distance from point $\boldsymbol{x}$ to the GMM is defined as the distance to the nearest cluster:

\begin{align}
	d(\boldsymbol{x}) = \min\limits_{k=1,\ldots,K} d_k(\boldsymbol{x}) \nonumber
\end{align}

Once the distance criterion is defined, we can establish a \emph{threshold} value
for new data as the maximum distance reached during the calibration period,
\emph{i.e.}:

\begin{align}
	d_\mathrm{th} = \max\limits_{j=1,\ldots,C-M} d(\boldsymbol{x}^j) \nonumber
\end{align}

During data processing, the distance to new points is evaluated, and if it
exceeds the established threshold $d_\mathrm{th}$ the region is considered as
an outlier. To avoid false positives and areas where the anomaly is detected
intermittently, moving-average distances are obtained over a sliding window of
size $S$ and compared to the threshold. Increasing $S$ provides a more reliable
criterion for declaring anomalies and avoiding false positives, although the
detection of true positives may be slightly delayed. A balanced value for $S$
is around 100 samples, which allows detection in approximately 1 second (for a
data rate of 100 samples/s), suitable for real-time CDPR operation. All scripts used for the implementation of the outlier detector in the CDPR setup are accessible via GitHub at: https://github.com/javiervales/cdpr. 

\subsection{Model update}
During the experimental phase (see Section~\ref{SETUP}~and~\ref{RESULTS}), data
drift was observed in torque measurements in many of the long running tests.
As previously mentioned, this may be due to external variations like
temperature or humidity, as well as internal effects related to the motors. For
example, this drift is clearly noticeable in
Figure~\ref{fig:test12} and can cause, as it does in that case,
normal regions to be erroneously treated as anomalies. Such misclassifications can be seen in the intermediate subplot of Figure~\ref{fig:test12}.

To avoid this problem, we introduced the following change: the algorithm
recalculates the GMM model when the mean distance of the sliding window of new
points exceeds $(1-\gamma)\times100$\% of the threshold but is less than 100\% of it, being
$\gamma$ a model hyper-parameter with a nominal value of $0.25$. A distance
in this range is a strong evidence that the GMM model has changed slightly and
should be readapted. The model is updated with the last $C$ anomaly-free registered samples. 
The output of this method is shown in the bottom subfigures of Figure~\ref{fig:test12and4}.
Now it correctly detects the zones with outliers through drift
correction. The instants when there has been a model update are
indicated with red inverted triangular marks in Figure~\ref{fig:test12and4}.

\section{Experimental results}
\label{RESULTS}

\begin{table*}[t!]
	\centering
	\caption{Percentage of True Positives (TP) and True Negatives (TN). Tests last 12 hours unless noted.}
		\resizebox{1\textwidth}{!}{%
			\begin{tabular}
				{m{0.75cm} L{9.5cm} R{1cm} R{1cm} R{1cm} R{1cm} R{1cm} R{1cm}}
				\toprule
				&   & \multicolumn{2}{l}{\celdat{Power method}} 
				& \multicolumn{2}{l}{\celdat{\begin{tabular}{@{}l@{}}Distance method\\(no updates)\end{tabular}}} 
				& \multicolumn{2}{l}{\celdat{\begin{tabular}{@{}l@{}}Distance method\\(updates)\end{tabular}}} \\
				\hline
				\celda{Test} & \celda{Description} & \celda{TP [\%]} & \celda{TN [\%]} 
				& \celda{TP [\%]} & \celda{TN [\%]} 
				& \celda{TP [\%]} & \celda{TN [\%]} 
				\\
				\hline
				\#1 & 
				\celdah{Type 1. 10/20\footnotemark min on/off cycles, wind strength 3} & 
				\celdah{0.0} & 
				\celdah{99.7} & 
				\celdah{100} & 
				\celdah{98.6} & 
				\celdah{100} & 
				\celdah{98.6} \\
				\hline
				
				\#2 & 
				\celdah{Steady conditions, no anomalies}  &
				\celdah{100} & 
				\celdah{5.0} & 
				\celdah{100} & 
				\celdah{15.1} & 
				\celdah{100} & 
				\celdah{100} \\
				\hline
				
				\#3 & 
				\celdah{Type 2. 10/20 min on/off cycles in the last 2 hours, wind strength 1} & 
				\celdah{0.0} & 
				\celdah{98.8} & 
				\celdah{100} & 
				\celdah{72.3} & 
				\celdah{100} & 
				\celdah{99.5} \\
				\hline
				
				\#4 & 
				\celdah{Type 1. 10/20 min on/off cycle, wind strength 2} & 
				\celdah{95.3} & 
				\celdah{4.8} & 
				\celdah{100} & 
				\celdah{58.3} & 
				\celdah{100} & 
				\celdah{97.9} \\
				\hline
				
				\#5 & 
				\celdah{Type 1. 1/2 hour on/off cycle, wind strength 1} & 
				\celdah{71.4} & 
				\celdah{24.3} & 
				\celdah{100} & 
				\celdah{17.1} & 
				\celdah{100} & 
				\celdah{100} \\
				\hline
				
				\#6 & 
				\celdah{Type 1. 1/2 hour on/off cycle, wind strength 3} & 
				\celdah{32.2} & 
				\celdah{17.1} & 
				\celdah{100} & 
				\celdah{36.9} & 
				\celdah{100} & 
				\celdah{99.4} \\
				\hline
				
				\#7 & 
				\celdah{Type 3. 6 hour test, variable wind strength in the last 5 hours} & 
				\celdah{58.1} & 
				\celdah{11.8} & 
				\celdah{100} & 
				\celdah{81.2} & 
				\celdah{100} & 
				\celdah{92.9} \\
				\hline
				
				\#8 & 
				\celdah{Type 3. 45 min test, anomalies in the last 15 min, wind strength 3} & 
				\celdah{0.0} & 
				\celdah{90.0} & 
				\celdah{100} & 
				\celdah{100} & 
				\celdah{100} & 
				\celdah{100} \\
				\hline
				
				\#9 & 
				\celdah{Type 1. 10/20 min on/off cycle, wind strength 1} & 
				\celdah{4.7} & 
				\celdah{95.2} & 
				\celdah{100} & 
				\celdah{85.2} & 
				\celdah{100} & 
				\celdah{100} \\
				\hline
				
				
				\#10 & 
				\celdah{Type 2. 2 hour test, 10/20 min on/off cycle in the last hour, wind strength 1} &        
				\celdah{24.2} & 
				\celdah{72.8} & 
				\celdah{100} & 
				\celdah{96.3} & 
				\celdah{100} & 
				\celdah{100} \\
				\hline
				
				\#11 & 
				\celdah{Type 2. 10 hour test, 10/20 min on/off cycle in the last 2 hours, wind strength 1} & 
				\celdah{27.3} & 
				\celdah{68.1} & 
				\celdah{100} & 
				\celdah{0.2} & 
				\celdah{100} & 
				\celdah{99.7} \\
				\hline
				
				\#12 & 
				\celdah{Type 2. 10/20 min on/off cycle in the last 2 hours, wind strength 3} & 
				\celdah{1.5} & 
				\celdah{99.2} & 
				\celdah{100} & 
				\celdah{18.7} & 
				\celdah{100} & 
				\celdah{99.4} \\
				\hline
				
				\#13 & 
				\celdah{Type 3. 20 hour test, variable strength anomalies in the last 10 hours} &  
				\celdah{0.0} & 
				\celdah{59.9} & 
				\celdah{100} & 
				\celdah{23.7} & 
				\celdah{100} & 
				\celdah{76.0} \\
				\hline

				\#14 & 
				\celdah{Type 3. 3 hour test, 20/30 min anomalies, alternate wind strengths 1 and 3} & 
				\celdah{0.0} &
				\celdah{95.5} & 
				\celdah{100} & 
				\celdah{50.0} & 
				\celdah{100} & 
				\celdah{72.7} \\
				\hline
				
				& 
				\celdah{\bfseries Average} & 
				\celdah{\bfseries 29.6} &
				\celdah{\bfseries 60.2} & 
				\celdah{\bfseries 100} & 
				\celdah{\bfseries 53.8} & 
				\celdah{\bfseries 100} & 
				\celdah{\bfseries 95.4} \\
				\hline
				\bottomrule
		\end{tabular} }
		\label{LATABLA}
	\end{table*}
	\footnotetext[1]{Format A/B: anomaly duration (A) and total period (B: wind + rest).}
	\linespread{1}
	
	This section describes the results obtained from applying the anomaly detection
	method in the experimental testbeds described in Section~\ref{SETUP}. 
	The primary objectives of the experimental phase were threefold:
	
	\begin{itemize}
		\item Validate improved strategies for anomaly detection.
		\item Study the phenomenon of temporal drift and how it affects anomaly detection.
		\item Establish parameterizations that provide robustness in de\-cision-making
		while allowing the algorithms to operate in real time. A target detection time
		of 1 second has been considered.
	\end{itemize}
	
	A total of 14 experiments were performed. Table~\ref{LATABLA} summarizes the configuration and results of each experiment and the previous Figure~\ref{fig:test12and4} showcased the results for three different methods:
	
	\begin{itemize}
		\item \textbf{Threshold method}: A threshold procedure applied to the moving
		average of the torque power signal. As in the distance method, the threshold is
		determined as the maximum obtained during the calibration phase.
		
		\item \textbf{Distance method without GMM updates}: The distance method
		described in the previous section, applied without updating the GMM.
		
		\item \textbf{Distance method with GMM updates}:
		The same distance method, but with GMM model updates incorporated.
	\end{itemize}
	
	In all cases, the moving average is computed over the last $S$=100 samples to meet
	the target detection time. For each of these methods, the percentages of true
	positives (TP) and true negatives (TN) are calculated: a TP occurs when an
	anomaly is correctly detected during the period when the fan is on; a TN occurs
	when no anomalies are detected during the period when the fan is off. These
	percentages are obtained by dividing the total time correctly classified as
	anomalous or non-anomalous by the actual total time with true anomalies (fan
	on) or without anomalies (fan off), respectively. Additionally, a guard period
	of 30 seconds is considered when the fan is turned off, since the CDPR should
	gradually return to the resting state. During this guard period, the detector's
	output is not included in the TP and TN statistics, as the system may produce
	ambiguous results while settling. For safety reasons, the system must never run in hazardous conditions. Every anomaly has to be identified, even if this means accepting a small number of false positives (occasionally flagging normal situations as anomalies).
	
	The experimental results demonstrate that the proposed distance method with GMM
	updates obtains consistent results across all experiments. Specifically, it
	achieved 100\% TP rates across all tests, effectively detecting all	anomalies. 
	It also reached very high TN rates, with an average of 95.4\%,
	indicating a low rate of false positives during normal operation. In contrast,
	the power threshold method showed poor performance, with an average TP rate of
	29.6\%. Regarding TN, the power and the distance model without updates only
	achieve rates in the order of 50\%, due to their inability to correct from
	operational drifts, making them unreliable for practical anomaly detection purposes.
	
	Figure~\ref{fig:additional_test} shows the results of the ``method with GMM updates'' on four additional experiments (\#6, \#7, \#9, and \#14). More specifically:
	
	\begin{figure*}[t!]
		\centering
		\begin{subfigure}[b]{0.49\columnwidth}
			\includegraphics[width=\linewidth]{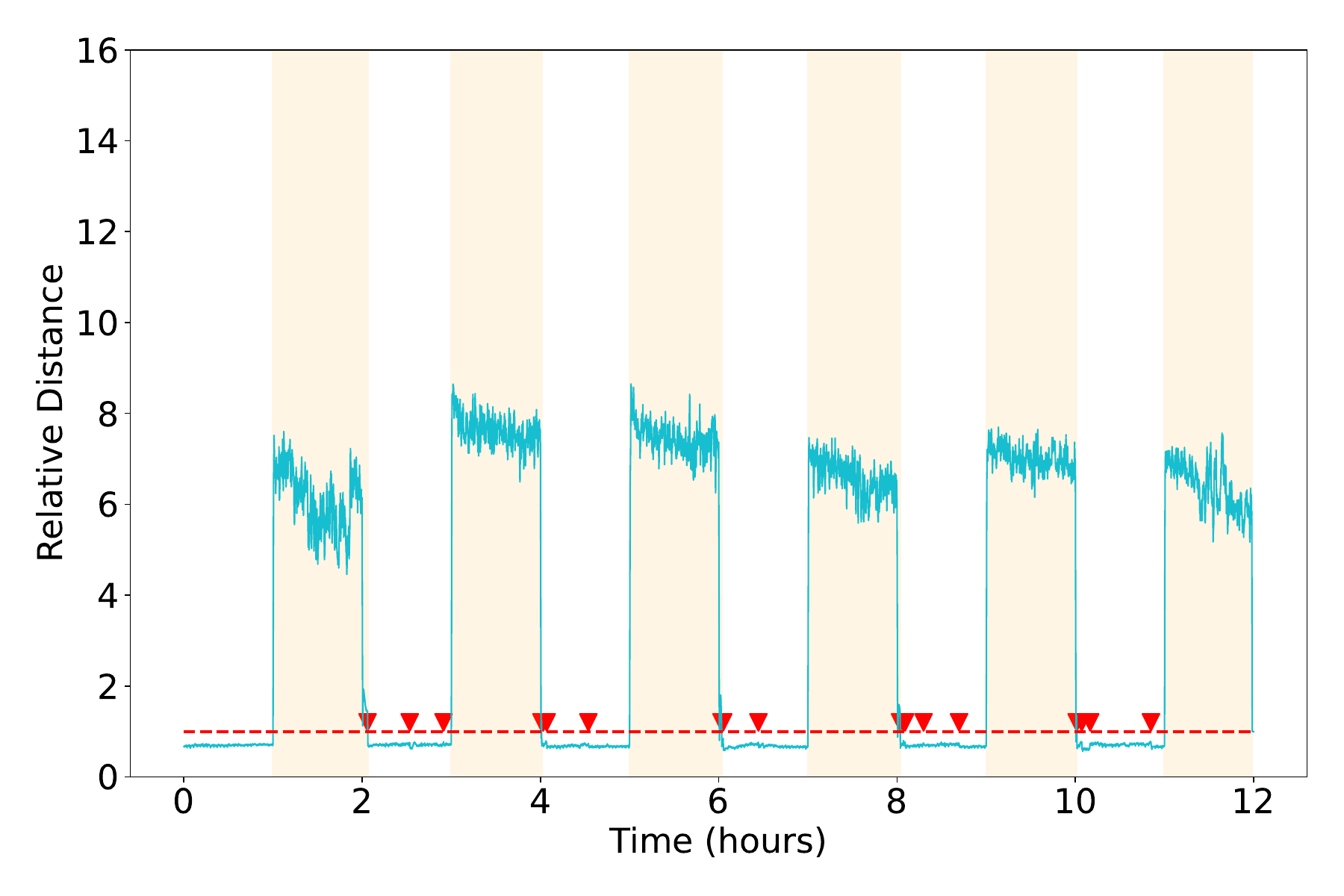}
			\caption{}
			\label{fig:test6}
		\end{subfigure}
		\hspace{0cm}
		\begin{subfigure}[b]{0.49\columnwidth}
			\includegraphics[width=\linewidth]{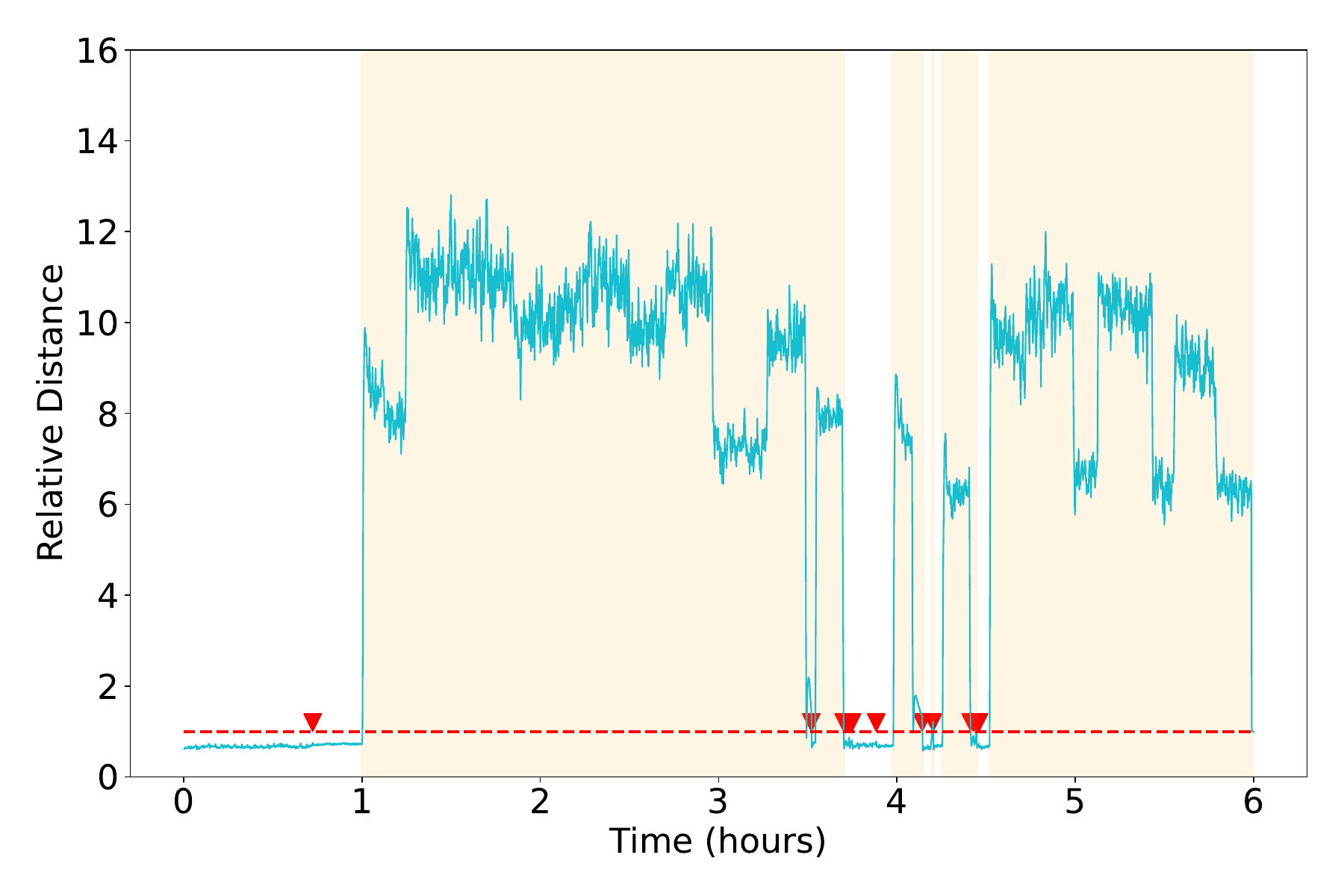}
			\caption{}
			\label{fig:test7}
		\end{subfigure}
		
		\begin{subfigure}[b]{0.49\columnwidth}
			\includegraphics[width=\linewidth]{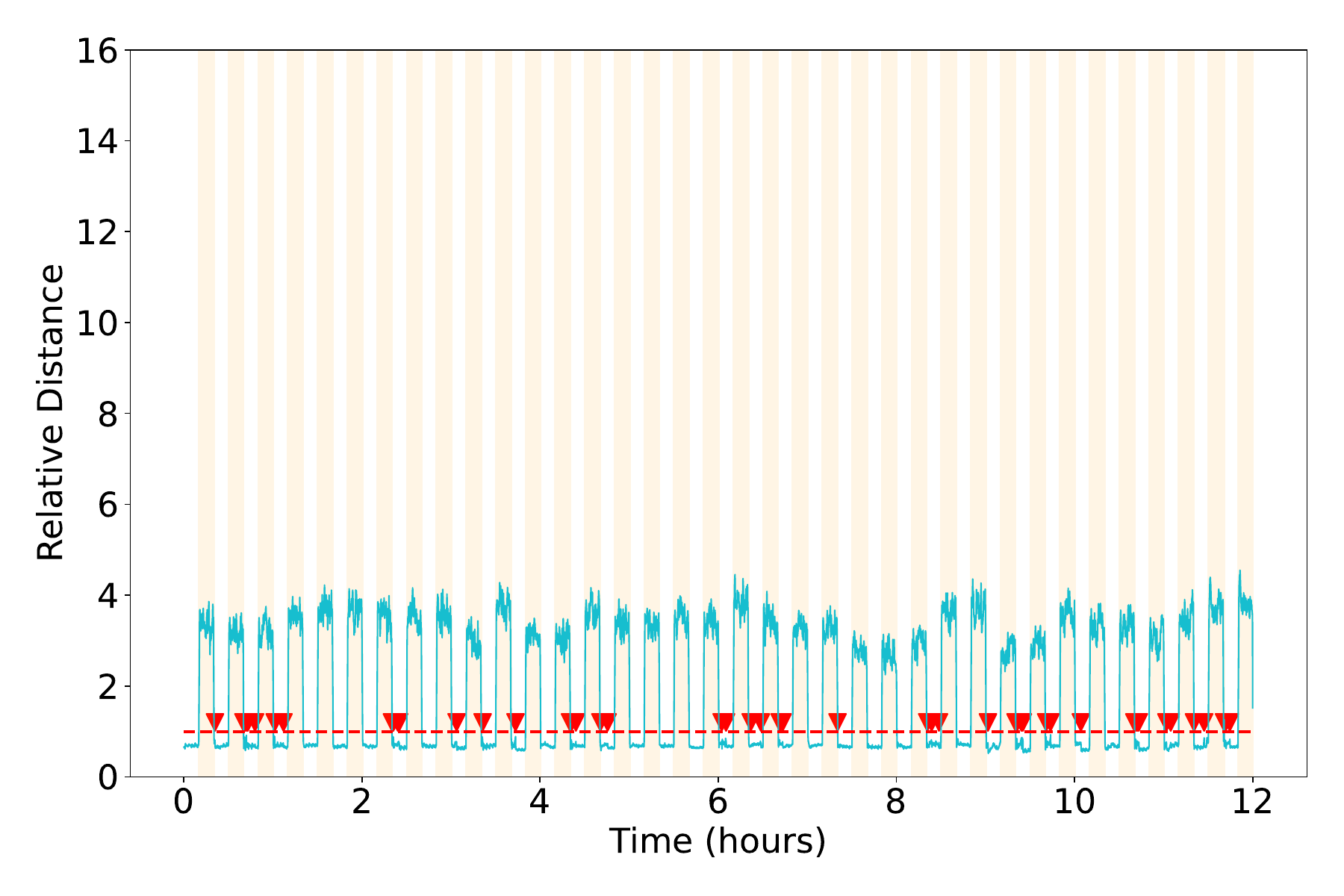}
			\caption{}
			\label{fig:test9}
		\end{subfigure}
		\hspace{0cm}
		\begin{subfigure}[b]{0.49\columnwidth}
			\includegraphics[width=\linewidth]{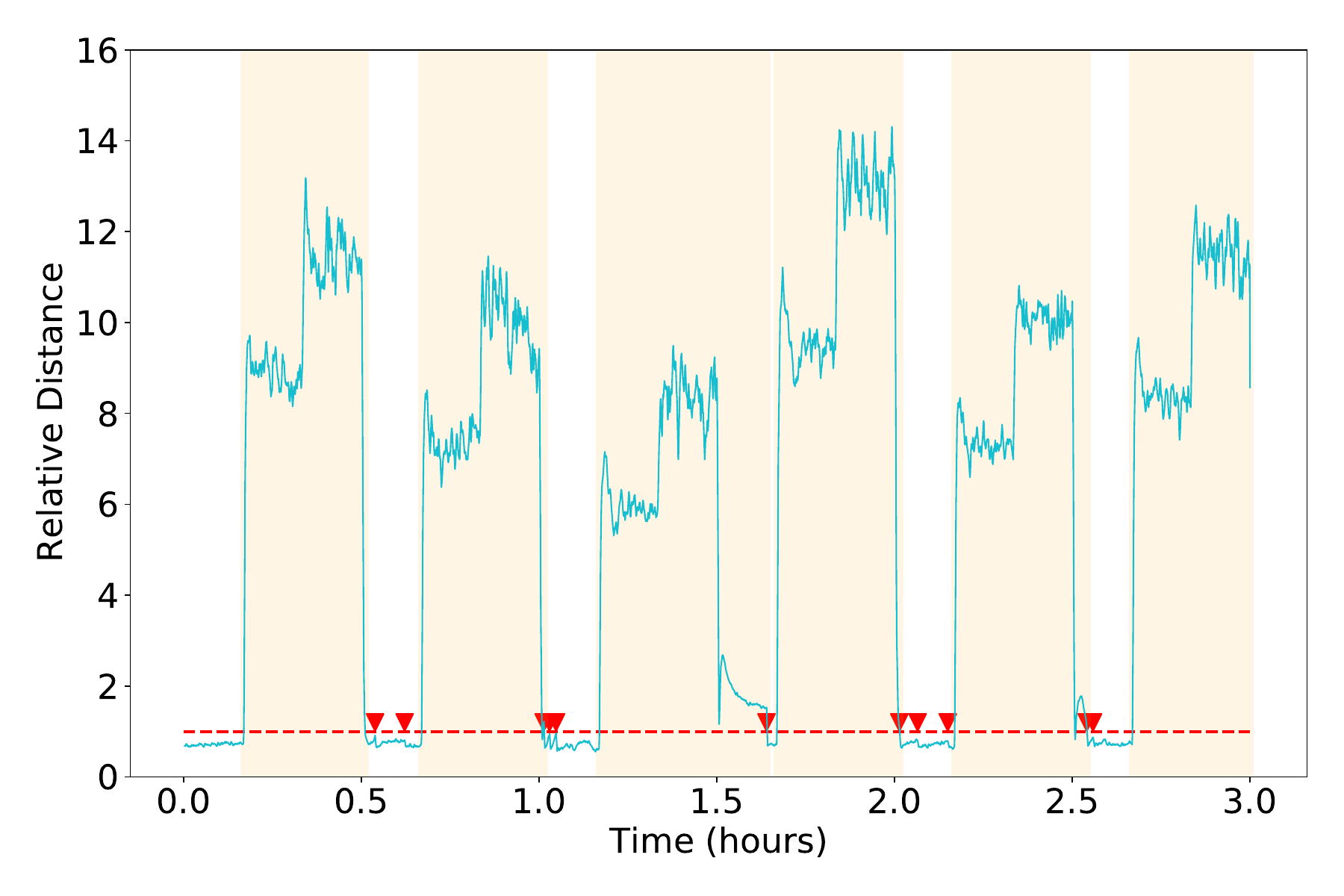}
			\caption{}
			\label{fig:test14}
		\end{subfigure}
		
		\caption{Moving average Mahalanobis Distance (model with updates) for experiments: (a) \#6, (b) \#7, (c) \#9, (d) \#14. For additional information see Table~\ref{LATABLA}.}
		\label{fig:additional_test}
	\end{figure*}
	
	\begin{itemize}
		\item Figure~\ref{fig:test6} shows the results of the experiment \#6 in which the CDPR alternates between one-hour rest and one-hour steady wind periods. During each calm period (e.g., at 3 h, 5 h, and 9 h) the method is updated and corrects drift. Also, this figure shows how the algorithm detects all anomalies above the threshold (yellow bands) and switches back to normal classification almost as soon as the anomaly ends, with a TN rate of 99.4\%. The few false positives (0.6\% of the samples) appear during the wind-to-rest transition. 
		\item Figure~\ref{fig:test7} shows the system under winds of varying intensity, as reflected by the variations in the relative distance signal. Although the signal fluctuates, it remains above the threshold, so the detector correctly identifies the whole period as anomalous. At about 3.5 h the wind stops, but the algorithm waits for a minimum settling time before it considers the system back to normal, so this brief interval is still labeled anomalous even when the distance suggests that it could be a nominal state. Additionally, there is a false positive at around 4.25 h, probably caused by the alterations in the model calculation. For safety reasons, the system flags an anomaly, and once the model is correctly updated, it returns to identify the true negatives accurately.
		\item In Figure~\ref{fig:test9}, all anomaly regions are correctly detected (100\% TP). When compared to the distance values in Figure~\ref{fig:test12and4}, a direct correlation between wind intensity and increasing distance is observed, which supports the expected behaviour of the system.
		\item In the test shown in Figure~\ref{fig:test14}, the algorithm perfectly identifies all outlier regions. However, because stability problems arise around 1.5 h, the model does not reach steady state until 1.64 h, which slightly delays accurate classification of the nominal states.	This represents the most challenging scenario tested, yet the system still meets the required specifications. It successfully identifies all anomalous events and maintains an acceptable accuracy in detecting nominal conditions (Table~\ref{LATABLA}). The delay in returning to normal classification is attributed to the temporary instability introduced by external winds, which delays model adjustments.
	\end{itemize}

\begin{figure}[t!]
		\centering
		\includegraphics[width=1.0\columnwidth]{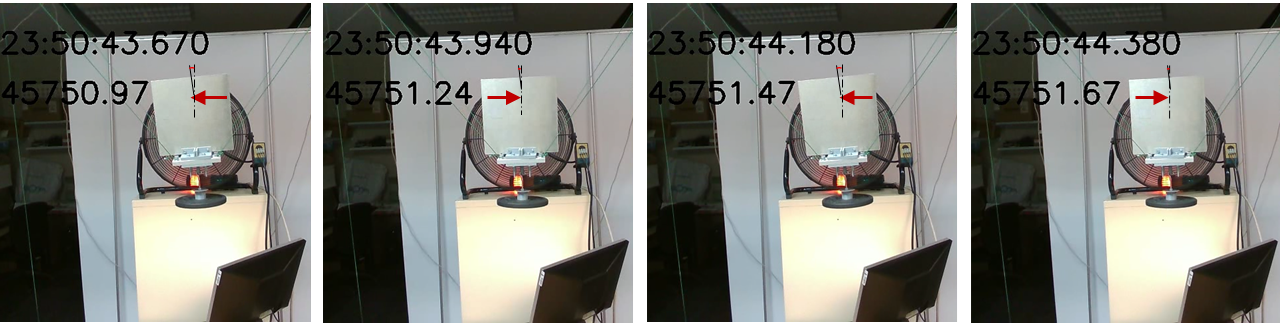}
		\caption{Example of detected anomaly. }
		\label{ANOMALIAEJEMPLO}
	\end{figure}

	As an example of anomaly detection, Figure~\ref{ANOMALIAEJEMPLO} illustrates the
	motion of the load when an anomaly was detected in test file \#14. This
	anomaly was caused by a maximum‐level (3) wind gust. Although the CDPR robot
	attempts to compensate for wind‐induced motion, a slight displacement of the
	payload remains perceptible (highlighted by the left and right arrows in the
	figure to facilitate identification of the movement). In other anomaly
	examples, the oscillatory movements are less visually evident due to the CDPR’s
	rapid compensation, yet they are still readily detected by our proposed
	algorithm through the resulting changes in the torques.
	
	\begin{figure*}[t]
		\centering
		\includegraphics[width=0.335\textwidth]{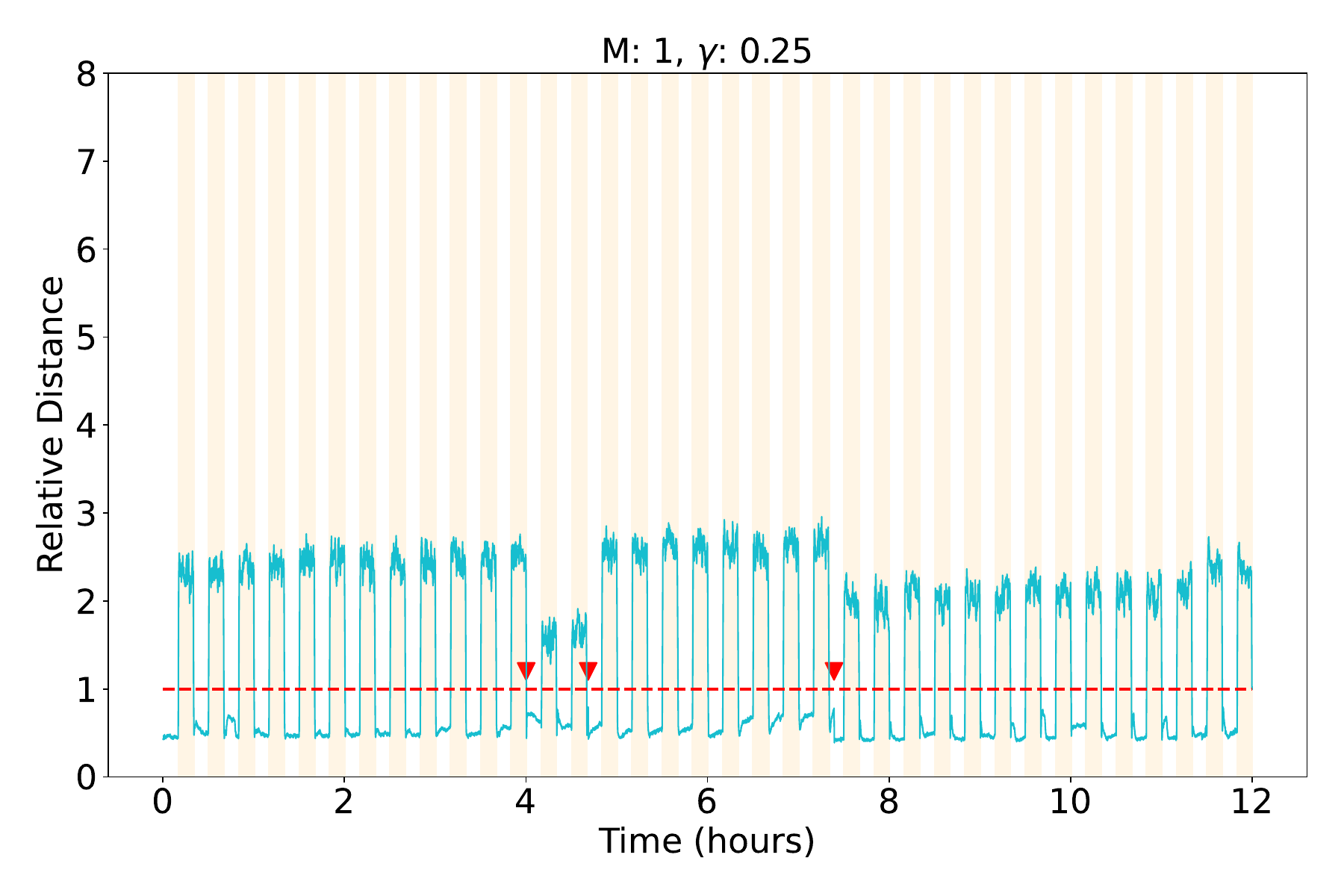}
		\includegraphics[width=0.335\textwidth]{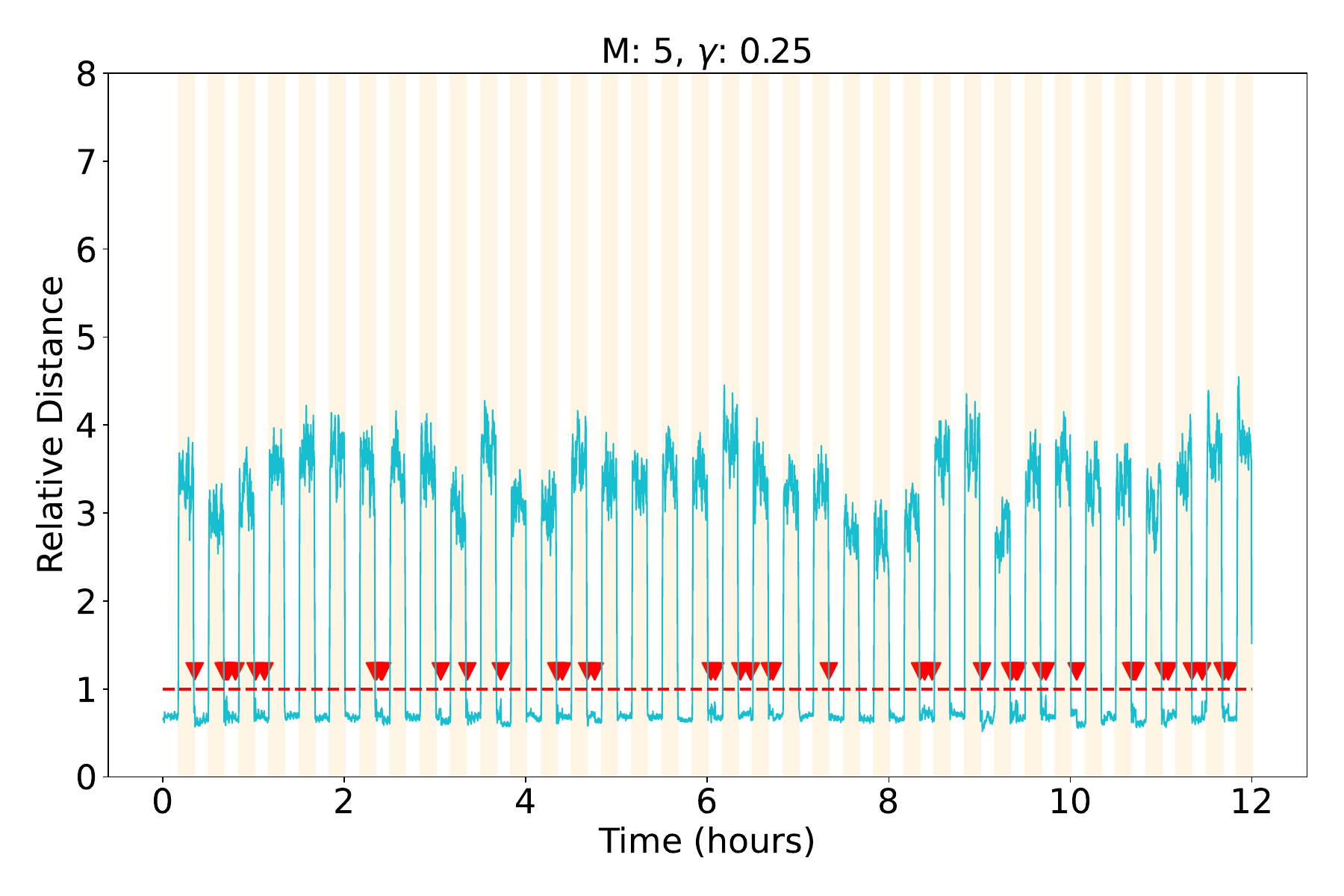}
		\includegraphics[width=0.335\textwidth]{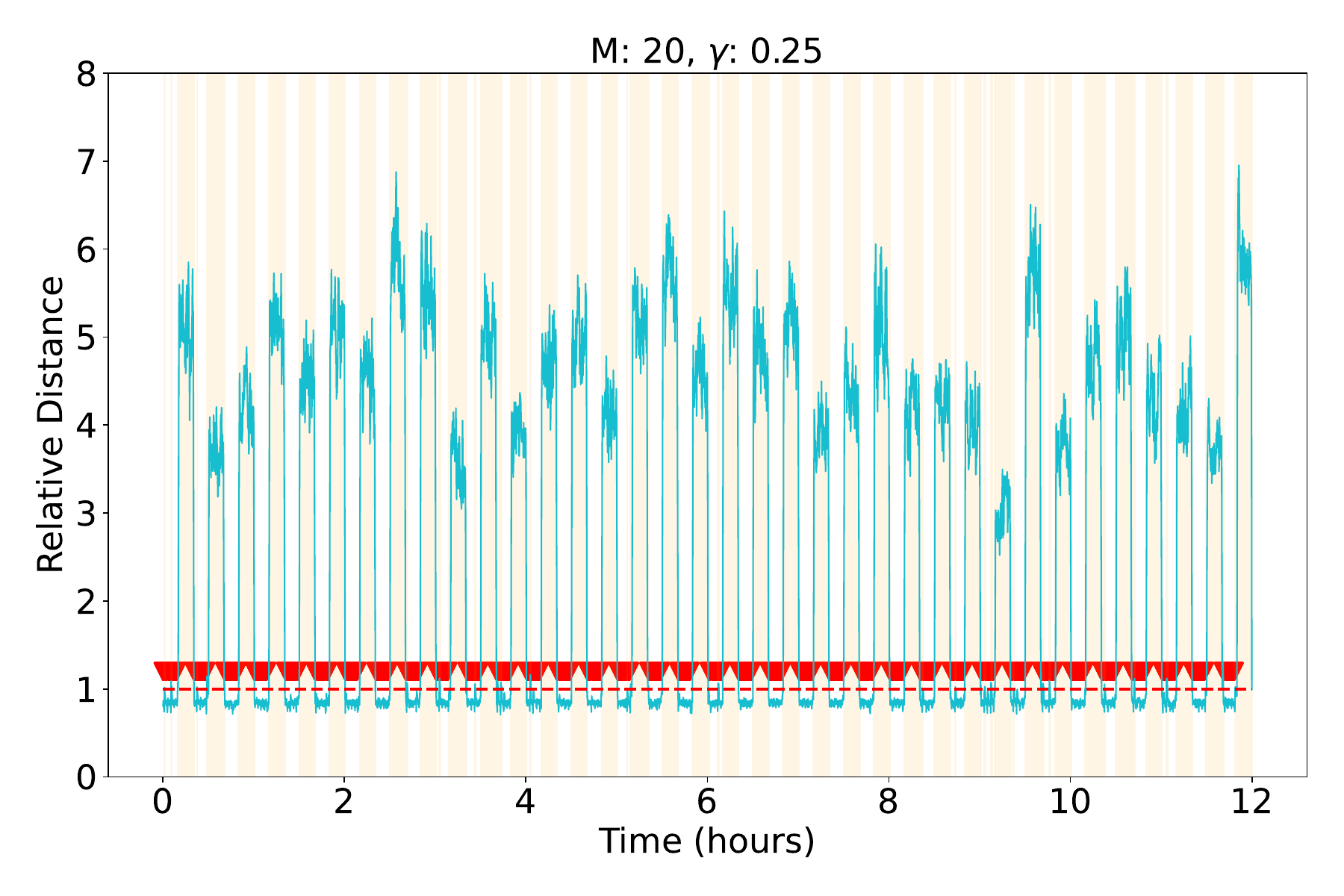}
		\caption{Effect of the model hyperparameter $M$ on the
			outcome of experiment \#9 (see Table~\ref{LATABLA}).}
		\label{HIPER}
	\end{figure*}
	
	Finally, the effect of the model's configuration hyperparameters was
	investigated. Modifications were tested on the model updating threshold
	$1-\gamma$. The tests did not show significant variations when setting this
	parameter between 0.5 and 0.95. Regarding the hyperparameter $M$, which
	controls the dimensionality of the model and was set at $M$=5 by default,
	various tests were conducted with values of $M$ from 1 to 20. Results were
	mostly satisfactory across the entire range of values although very low values
	result in a reduced distance dynamic range, which may, in rare occasions, lead
	to misclassifications. Conversely, when dimensionality is too high, the model
	becomes unstable and constant updates are necessary, which could affect
	computational performance, as shown in Figure~\ref{HIPER}.
	
	\section{Conclusion}
	\label{CONCLUSIONS}
	In this paper, we presented an unsupervised anomaly detection algorithm for CDPRs while the platform is at rest, relying solely on motor-torque measurements and requiring no extra sensors. This anomaly detection is particularly relevant for CDPR operating modes in which the end-effector must hold a fixed pose while motor torque remains active, as is commonly demanded during machining operations and grip/release steps in pick-and-place tasks.
	
	Experimental results demonstrated that the proposed ``method with GMM updates'' efficiently detects anomalies and adapts to temporal drifts, maintaining high sensitivity while minimizing false positives. When compared to the relative power method and the GMM method without updates, our method consistently outperformed both in the face of drift and changing environments, all without requiring any labeled anomaly data. Across fourteen experiments the proposed method achieved an average 100\% detection rate and correctly classified 95.4\% of normal conditions, with every anomaly detected in one second.
	
	Its simplicity and minimal parameter tuning make it well-suited for real-time applications, providing a practical solution to enhance the safety of CDPR operations. Future work will focus on extending the algorithm to other operational phases of CDPRs.

\textbf{Acknowledgements} \par 
Diego Silva-Mu\~niz is grateful to the Universidade de Vigo for his postdoctoral orientation grant (00VI 131H 6410211-2023).

	\medskip
	\textbf{Author Contributions} \par 
	The study conception and design were carried out by Julio Garrido and Javier Vales. Material preparation, data curation, and investigation were conducted by Julio Garrido, Javier Vales, Diego Silva-Muñiz, and Enrique Riveiro. Formal analysis was performed by Julio Garrido and Javier Vales. Software development was carried out by Javier Vales, Josu\'e Rivera and Enrique Riveiro, while visualization was handled by Josu\'e Rivera and Pablo L\'opez-Matencio. The first draft of the manuscript was written by Julio Garrido, Javier Vales, Diego Silva-Muñiz and Enrique Riveiro, while writing review and editing were performed by Julio Garrido, Javier Vales, Diego Silva-Muñiz, Pablo L\'opez-Matencio and Josu\'e Rivera. All authors read and approved the final manuscript.
	
	\medskip
	\textbf{Conflict of interest} \par 
	The authors declare that they have no known competing financial interests or personal relationships that could have appeared to influence the work reported in this paper.
	
	\medskip
	\textbf{Data Availability Statement} \par 
	All sources and csv datafiles are available at GitHub: https://github.com/javiervales/cdpr.

\bibliographystyle{unsrt}
\bibliography{GMM_anomaly_detector_CDPR}  






\end{document}